# COVID-19 cases prediction using regression and novel SSM model for non-converged countries

Rupali Patil[a,*], Umang Patel[b], Tushar Sarkar[c]

[a]*KJ Somaiya College of Engineering,Mumbai-400077, India, rupalipatil@somaiya.edu*
[b]*KJ Somaiya College of Engineering,Mumbai-400077, India, umang@somaiya.edu*
[c]*KJ Somaiya College of Engineering,Mumbai-400077, India, tushar.sarkar@somaiya.edu*

**Abstract**

Anticipating the quantity of new associated or affirmed cases with novel coronavirus ailment 2019 (COVID-19) is critical in the counteraction and control of the COVID-19 flare-up. The new associated cases with COVID-19 information were gathered from 20 January 2020 to 21 July 2020. We filtered out the countries which are converging and used those for training the network. We utilized the SARIMAX, Linear regression model to anticipate new suspected COVID-19 cases for the countries which did not converge yet. We predict the curve of non-converged countries with the help of proposed Statistical SARIMAX model (SSM). We present new information investigation-based forecast results that can assist governments with planning their future activities and help clinical administrations to be more ready for what's to come. Our framework can foresee peak corona cases with an R-Squared value of 0.986 utilizing linear regression and fall of this pandemic at various levels for countries like India, US, and Brazil. We found that considering more countries for training degrades the prediction process as constraints vary from nation to nation. Thus, we expect that the outcomes referenced in this work will help individuals to better understand the possibilities of this pandemic.

## 1. Introduction

Currently, people all over the world have been affected by coronavirus disease 2019 (COVID-19), which is the fifth pandemic after the 1918 flu pandemic [1]. COVID-19 emerged in the city of Wuhan, China in late December 2019 and was declared a global pandemic by the World Health Organization (WHO) on 11 March 2020[2] Certain irresistible diseases occur in exponential or bell shape patterns. The countries which have not yet reached its peak value yet and shows exponential growth are referred as non-converged countries. The study of such patterns results in useful information for the prediction of new cases. Because of this, it is conceivable to find a way to limit its effect. Social media platforms were also used for such predictions. Lei Qin *et.al.* collected Social media search indexes (SMSI) for dry cough, fever, chest distress, coronavirus, and pneumonia from 31 December 2019 to 9 February 2020. Five methods, namely subset selection, forward selection, lasso regression, ridge regression, and elastic net, were used to predict new suspected COVID-19[3]. Petropoulos *et. al*. [4] concentrate on the combined daily figures totalled universally of the three main factors: confirmed cases, deaths, and recoveries. This paper depicts the course of events of a live predicting exercise and gives target estimates to the affirmed instances of COVID-19. Li *et.al*. performed forward and backward analysis for propagation analysis and prediction of the COVID-19. Existing data of the Hubei epidemic is used for the predictions of the epidemic development trends in South Korea, Italy, and Iran [5]. The Covid-19 pandemic is a significant worldwide emergency since the Second World War. Surpassing the size and scope of the repercussions of a World War, it has influenced all the Countries of our planet. To date, the quantity of Covid-19 deaths surpasses 75,000 and is tragically bound to exponentially develop in the coming weeks [6]. Chakraborty et al. used a two-fold approach. The first approach was for predication of new COVID-19 cases for various countries using hybrid autoregressive integrated moving average-wavelet based forecasting model. Another approach was for risk assessment for some intensely affected countries using the optimal regression tree algorithm [7].

* Corresponding author.
*E-mail address*: rupalipatil@somaiya.edu (Rupali Patil)

In the year 2014, Kane *et al*. [8] worked in the same field. He predicted the outbreak of avian inuenza H5N1 in the nearest future. Kane et al. used ARIMA and Random Forest time series models for predicting this outbreak. In the study, Kane et al. found that Random Forest time-series predictive analysis provided better results over the ARIMA model. Liu et. Al. estimated epidemiological parameters in the models, such as the transmission rate and the basic reproductive number, using data of infected persons [9]. Hamzah *et al* developed an online platform (Corona Tracker) [10] that provides the latest information, analysis, and development related to COVID-19. Corona Tracker assists with interpreting examples of individual assumption related to wellbeing data, and evaluate the political and financial impact of the spread of the infection. An understanding of the COVID-19 spreading status in cities can be used for pattern analysis, crisis management, medical services, supply of pharmaceutical-food materials, and specialized forces transportations, etc [11]

The colossal mortality, illness, and monetary log jam because of current COVID-19 pandemic has carried practically all nations to their knee. Current World populace is 7.77 billion and with a populace of 1.37 billion, India is the second-most crowded nation. India has stood apart of the group in its way to deal with COVID-19, thus far, has effectively contained the network transmission in many districts of India. As on 26 July 2020, 08:00 IST there were 467,882 Active Cases, 885,576 Cured/Discharged, 1 Migrated and 32,063 Deaths in India. This is the most elevated ever recuperation registers in a solitary day with more than 36,000 patients releases. With another high recuperation pace of 64% according to the reports of the world health organization. Genetic Programming model was used by Salgotra [12] for time series analysis and forecasting of the COVID-19 Pandemic in India. The highest accuracy to date was achieved by Salgotra reported 99.99 % but only for a region of Delhi and their accuracy in other regions has gone down below 98.8 %. Roosa [13] used the forecasting model for the cumulative number of confirmed reported cases in Hubei province in China from February 5th to February 24th, 2020 and could reach only 98 % accuracy. Dhanwant *et al*. [14] worked in the equivalent field for foreseeing the spread of Coronavirus in various nations by considering different highlights of Coronavirus development. They found that the factor $\beta$ administered by different elements like the social-contact structure. The reliance of this factor $\beta$ on the transmission level in the public arena gives a feeling of the viability of the measures taken for social distancing. The study by Pandey *et.al*. showed outbreak and analysis of COVID-19 for India till 30th March 2020 and predictions have been made for the number of cases for the next 2 weeks [15]. They have estimated that cases may rise between 5000-6000 in the next two weeks of time. The regional analysis of Indian states and the preparedness level of India in combating this outbreak was discussed by Gupta *et.al.* The major findings show that number of infected cases in India is rising quickly with the average infected cases per day rising from 10 to 73 from the first case to the 300th case [16]. Ranjan *et.al*. proposed exponential and classic susceptible-infected-recovered (SIR) models [17] based on available data that are used to make short and long-term predictions on a daily basis. Based on the SIR model, it is estimated that India will enter equilibrium by the end of May 2020 with the final epidemic size of approximately 13,000.

## 2. Materials and Methods

### *2.1 Data collection*

The prevalence of data was obtained from worldometer website (https://www.worldometers.info/coronavirus/).

Microsoft Excel was used for further analysis of data for 207 countries between 22 January 2020 to 21 July 2020. The features available in the dataset were country, continent, population, the total number of cases. Total deaths, recovered, and active cases were also there. Figure 1 shows the flow diagram for the prediction of the number of COVID-19 infected cases. Cumulative data from 207 countries is pre-processed and filtered for day wise and country wise analysis. The linear regression model then finds peak day and peak value for infected cases for countries filtered out for three different conditions. The peak cases, peak days and average of selected countries were used as input to Statistical SARIMAX model (SSM) for prediction of peak and fall down of number of cases for COVID-19 for countries like India, US, and Brazil.

### *2.2 Pre-processing, filtering and averaging*

The available database is for the duration of 22 January 2020 to 21 July 2020. Pre-processing was done for day-wise sorting and country-wise sorting is done by filtering the given database. Here country-wise filtering is done for three different conditions. In the first step of filtering condition used was, the population more than 20 million and peak days less than140 days and total cases per one million population greater than 500, for a selection of 12 countries. For the selection of 20 countries condition used was population more than 10 million and peak days less than140 days and total cases per one million population greater than 400. Population more than 5.5 million and the number of days less than150 days and total cases per one million population greater than 100 was used for selection of 41 countries in the third step of filtering. These three different filtering approaches were then fed to the linear regression model.

### *2.3 Linear regression model*

A linear regression model is used to predict the numerical value of *Y* for a given value of *X* using a straight line (called the *regression* line). If you know the slope and the y-intercept of that regression line, then you can plug in a value for X and predict the value for Y. In other words, you predict Y from X. Here regression model is used for prediction of maximum (peak) number of infected cases and the day at which this peak arrives. The input features for linear regression model are population, total number of cases per 1 million population and a population density which provides output features as total cases, peak values, and peak days for three filter conditions.

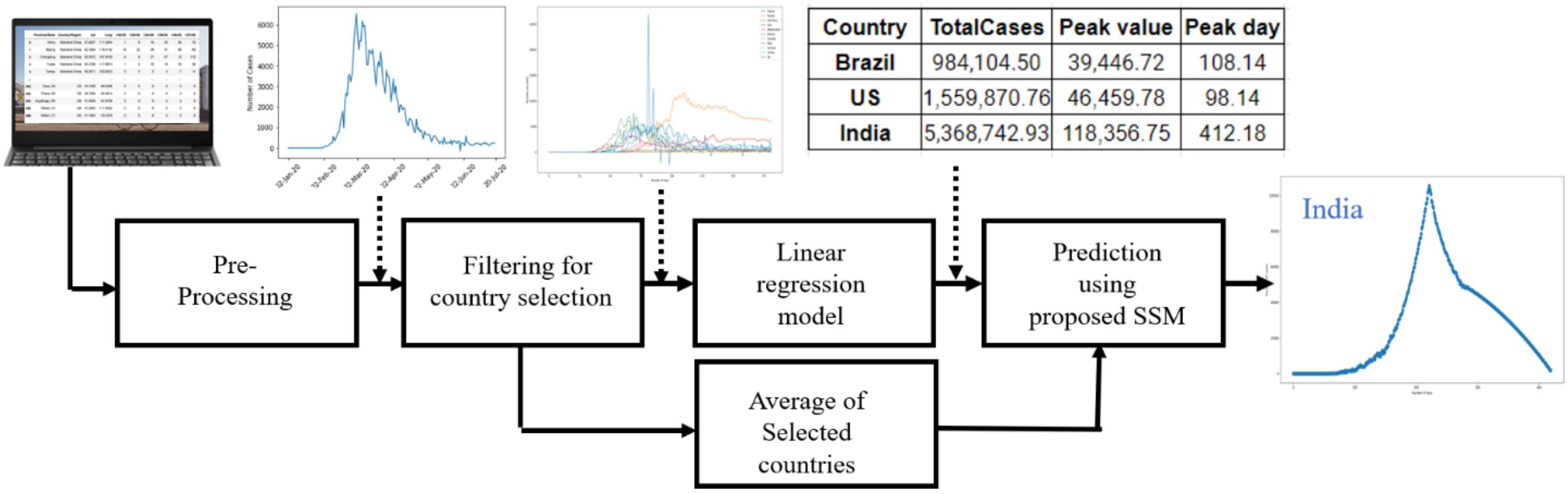

**Fig. 1.** Flow diagram for prediction of number of covid-19 infected cases. Cumulative data from various countries is pre-processed and filtered for day wise and country wise analysis. Linear regression model then finds peak day and peak value for infected cases. The peak cases, peak days and average of selected countries were used as input to Statistical SARIMAX model (SSM) for prediction of peak and fall down of number of cases for covid-19.

### *2.4 Proposed Statistical SARIMAX model (SSM)*

With the help of this model, we predicted the peak value as the increase in the number of cases per day. Then all the values up to this peak value were predicted with the help of the SARIMAX model which was trained on available data. After analyzing the graphs of many countries we came to the conclusion that the variance of the graph after reaching its peak value was more than it was before reaching its peak value So to extract more information from this we calculated the ratio of the mean values before and after the peak value for each country and took its average to predict the mean for our model as shown in equation 1. This average was added to all the values to get the prediction about the convergence of the curve and estimated number of days as per equation 4.To remove the saturation which aroused due to the mean value we used statistical SARIMAX model(SSM) trained over the 400 days values and got the predicted outputs graph.

$$Mean = \frac{1}{N}\sum_{i=0}^{N}\frac{R_i}{F_i} \quad (1)$$

$$where\ R_i = \frac{1}{p}\sum_{0}^{p} x_i \quad (2)$$

$$F_i = \frac{1}{n-p}\sum_{p}^{n} x_i \quad (3)$$

$$Bias = \frac{mean\ of\ rising\ edge\ of\ predicated\ country}{Mean} \quad (4)$$

Where N was no of selected Countries based on filter, p was no of days requires to reach peak value, n was total no. of days, $R_i$ was mean of $i^{th}$ country for rising edge of curve and $F_i$ was mean of $i^{th}$ country for falling edge of curve.

## 3. Results and Discussions

The raw data which is available on worldometer website required some pre-processing to make it suitable for our model. We handled some missing values by using the right imputation based on its other features. After this, we tried to extract the features which affected our model the most by proper exploratory data analysis and selected those features as the input for the final model. Figure 2 (a) shows that the cumulative prevalence of COVID-2019 presented an increasing trend that is reaching the epidemic plateau for countries like Italy, Iran, Russia etc.

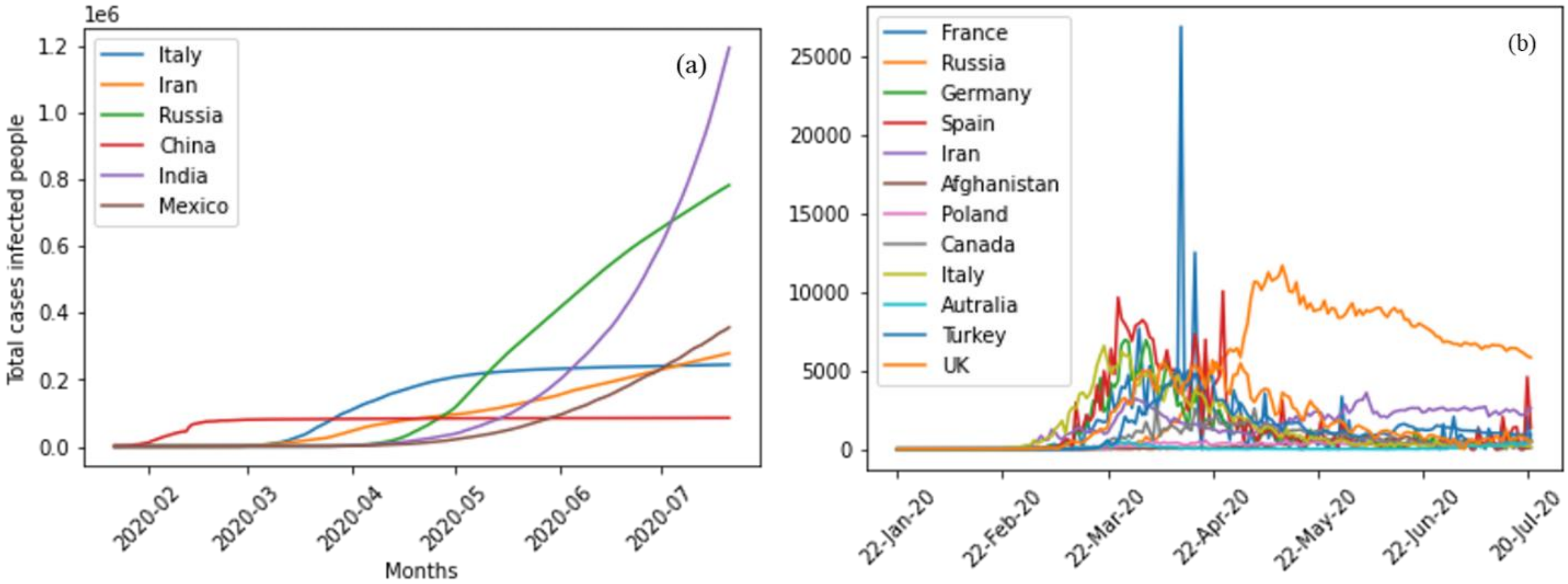

**Fig 2** (a) The trend of infections in some countries, (b) The graph of corona infections per day of the 12 countries chosen by fliter1.

As we can observe that the R-squared value decreases as the number of countries are increasing as per filter 2 and 3, which shows that the performance of the model goes down as the similarity of the training data with the target data decreases. The linear regressive model with all three filters is used for the final prediction for all non-converged countries and this can be seen in figure 3 (c). The regressive model also predicts the total number of infected people and by using different filters we get different predictions. Also, the comparison of the predicted peak values and the total number of predicted cases of the target countries India, US, and Brazil is shown in figure 3 (a) and (b) for further insight for filters 1,2, and 3.

Figure 4 shows the different filter conditions and their effects on the final result. The first condition was used for getting the countries that have the most similar features with the target countries. The target countries have a large population and so we had to set the conditions after getting such insights. We got 12 countries after setting the first condition. Our model was trained on these and the last two columns show its result. The peak value comes out to be 1.05 lakhs and the time required to gain its peak value is 221 days. We got an R-squared score of 0.986 on training these values. The second condition was used for getting some more countries which have less similarity than the target countries. We obtained 20 countries after setting this condition. Our model was again trained on these values to get an R-squared score of 0.9235. The results of the second model show that the peak value is 1.19 lakhs with a time of 227 days to gain this value. We further increased the number of countries by decreasing the similarity by using the third condition. We took 41 countries for training and got r squared score as 0.7798. As we can see in its average graph that it differs significantly from the other graphs. The peak value for this case was 1.31 lakhs and it required 232 days to reach that value. We can predict any non-converged countries from this model. We had selected the most affected countries for results which were India, US, and Brazil.

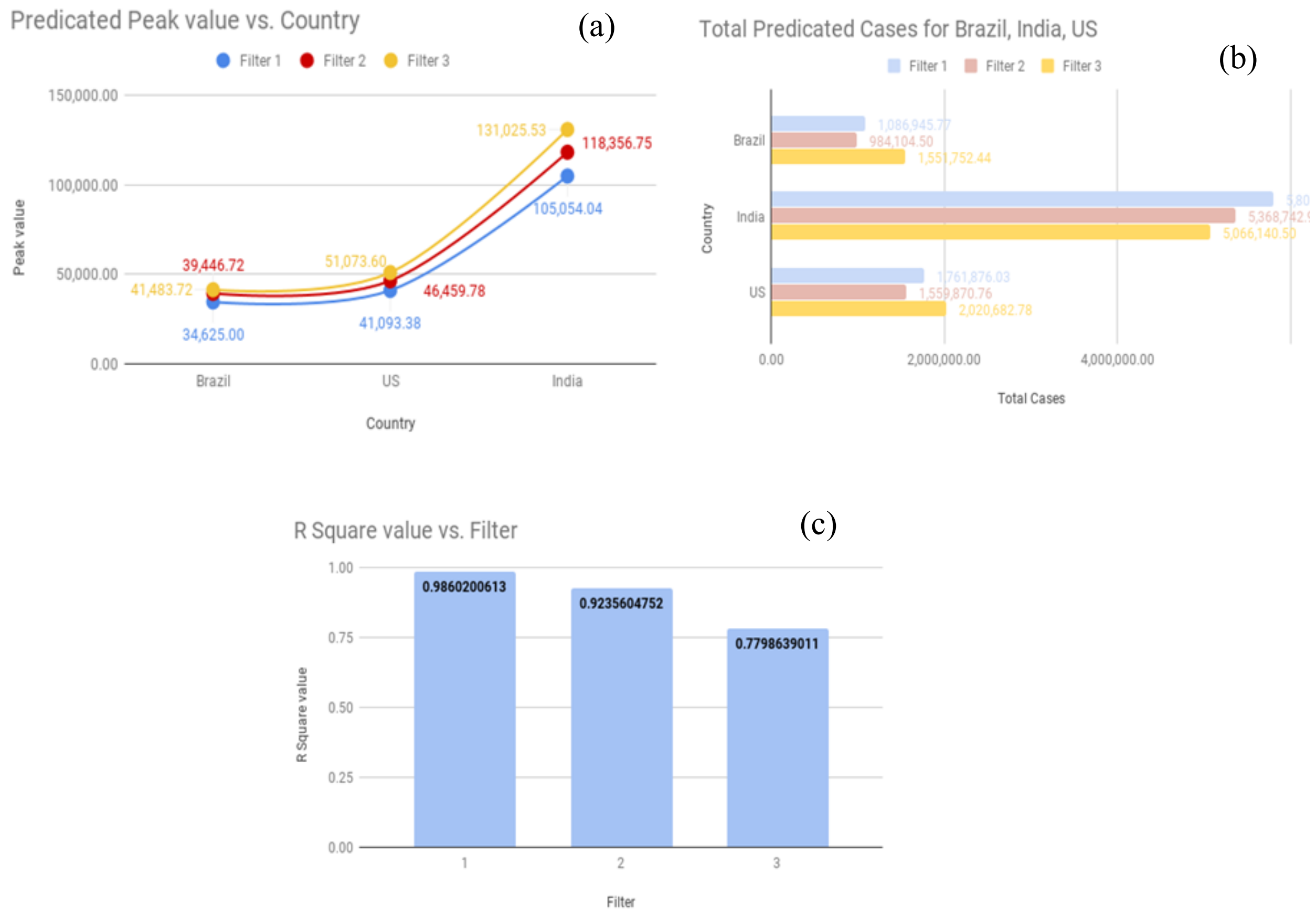

**Fig. 3** (a) The predicted peak values for three countries using the filter 1,2 and 3 (b) The total predicted cases of the target countries for each filter. (c) The R-squared value of the countries in each filter.

<table>
<tr><th>Filtering for country selection</th><th>Predicated by Linear regression model</th><th>Average of Selected countries</th><th>Prediction using Proposed SSM (India)</th></tr>
<tr><td>Filter 1=data['Tot Cases/1M pop'] > 500) & (data['Peak_days'] < 140) & (data['Population'] > 20000000</td><td><table><tr><th>Country</th><th>Total Cases</th><th>Peak value</th><th>Peak day</th></tr><tr><td>Brazil</td><td>1,086,945.77</td><td>34,625.00</td><td>134.98</td></tr><tr><td>US</td><td>1,761,876.03</td><td>41,093.38</td><td>126.37</td></tr><tr><td>India</td><td>5,804,118.74</td><td>105,054.04</td><td>535.92</td></tr></table></td><td>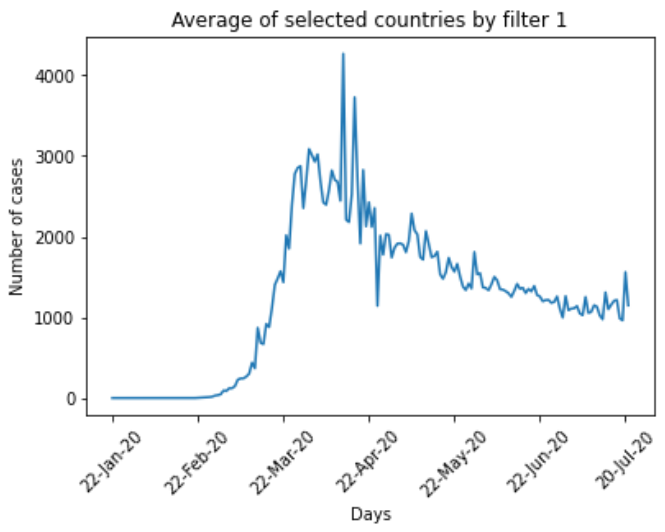
</td><td>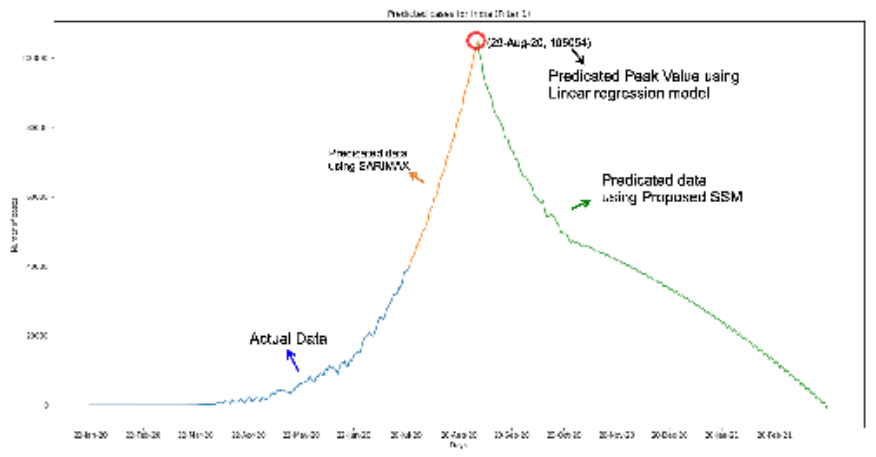
</td></tr>
<tr><td>Filter 2=(data['Tot Cases/1M pop'] > 400) & (data['Peak_days'] < 140) & (data['Population'] > 10000000 )</td><td><table><tr><th>Country</th><th>Total Cases</th><th>Peak value</th><th>Peak day</th></tr><tr><td>Brazil</td><td>984,104.50</td><td>39,446.72</td><td>108.14</td></tr><tr><td>US</td><td>1,559,870.76</td><td>46,459.78</td><td>98.14</td></tr><tr><td>India</td><td>5,368,742.93</td><td>118,356.75</td><td>412.18</td></tr></table></td><td>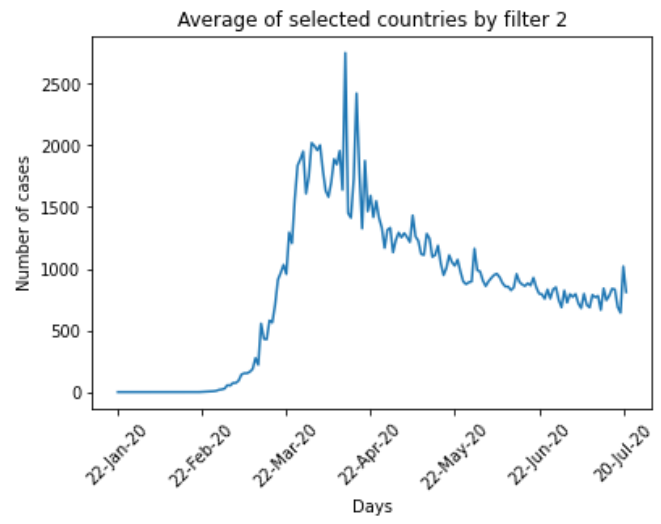
</td><td>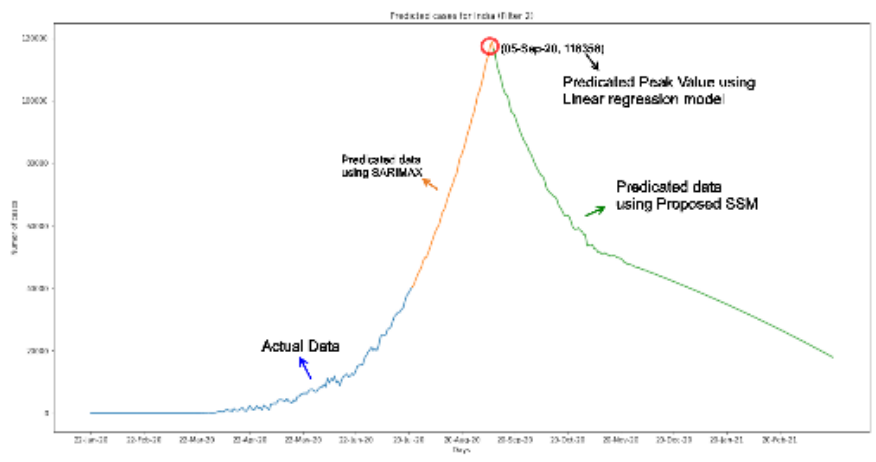
</td></tr>
<tr><td>Filter 3=(data['Tot Cases/1M pop'] > 100) & (data['Peak_days'] < 150) & (data['Population'] > 5500000 )</td><td><table><tr><th>Country</th><th>Total Cases</th><th>Peak value</th><th>Peak day</th></tr><tr><td>Brazil</td><td>1,551,752.44</td><td>41,483.72</td><td>150.22</td></tr><tr><td>US</td><td>2,020,682.78</td><td>51,073.60</td><td>136.60</td></tr><tr><td>India</td><td>5,066,140.50</td><td>131,025.53</td><td>323.82</td></tr></table></td><td>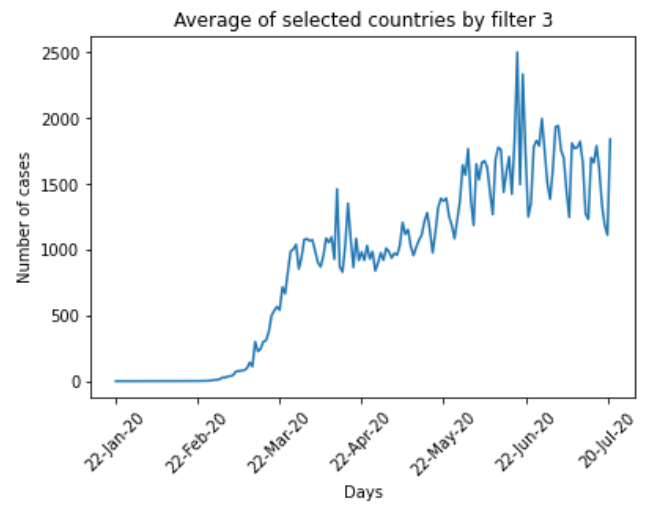
</td><td>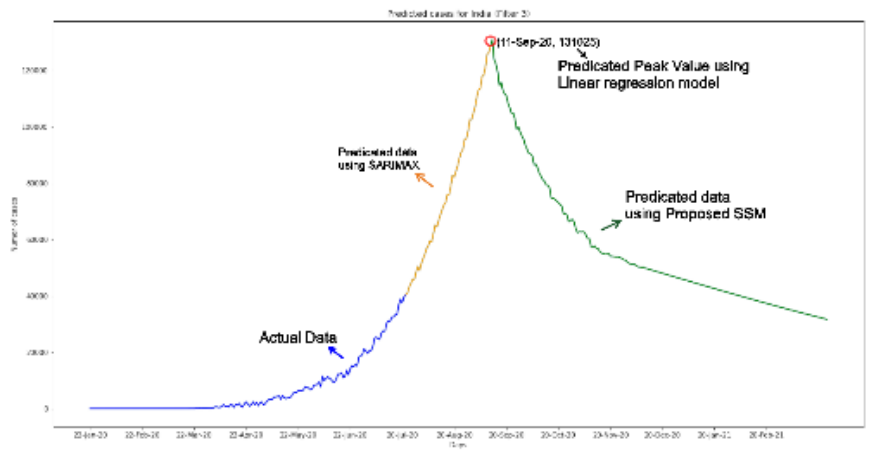
</td></tr>
</table>

**Fig 4** Column 1 shows the condition used for the filters, Column 2 shows the predictions done by the linear regression model, Column 3 shows the average of the countries selected by the filter and Column 4 shows the final graph of India with the help of proposed SSM model.

The predicted curve for India is determined by using the proposed SSM model on the countries selected by the filters. These three curves are then combined to give figure 5. We calculated the average of the predictions for the filters. These average predictions are then used for plotting figure 5.

## 4. Conclusion

We initially selected countries based on filtering concepts which are converged and used as train the model. Estimate the number of Total cases and Peak value using a linear regression model for non-converge countries. The peak value will be observed in the last week of August or the first week of September and the cases per day will range between 1.05 lakhs to 1.31 lakh for country India. The number of countries selected to train the model is increased based on the different filters. We observe that the performance of the model decreases with an increase in the number of countries. We estimated the curve of India by using the Proposed SSM model. We used linear regression, SARIMAX model, and proposed SSM model to get the effect of hybrid model which provided better accuracy.

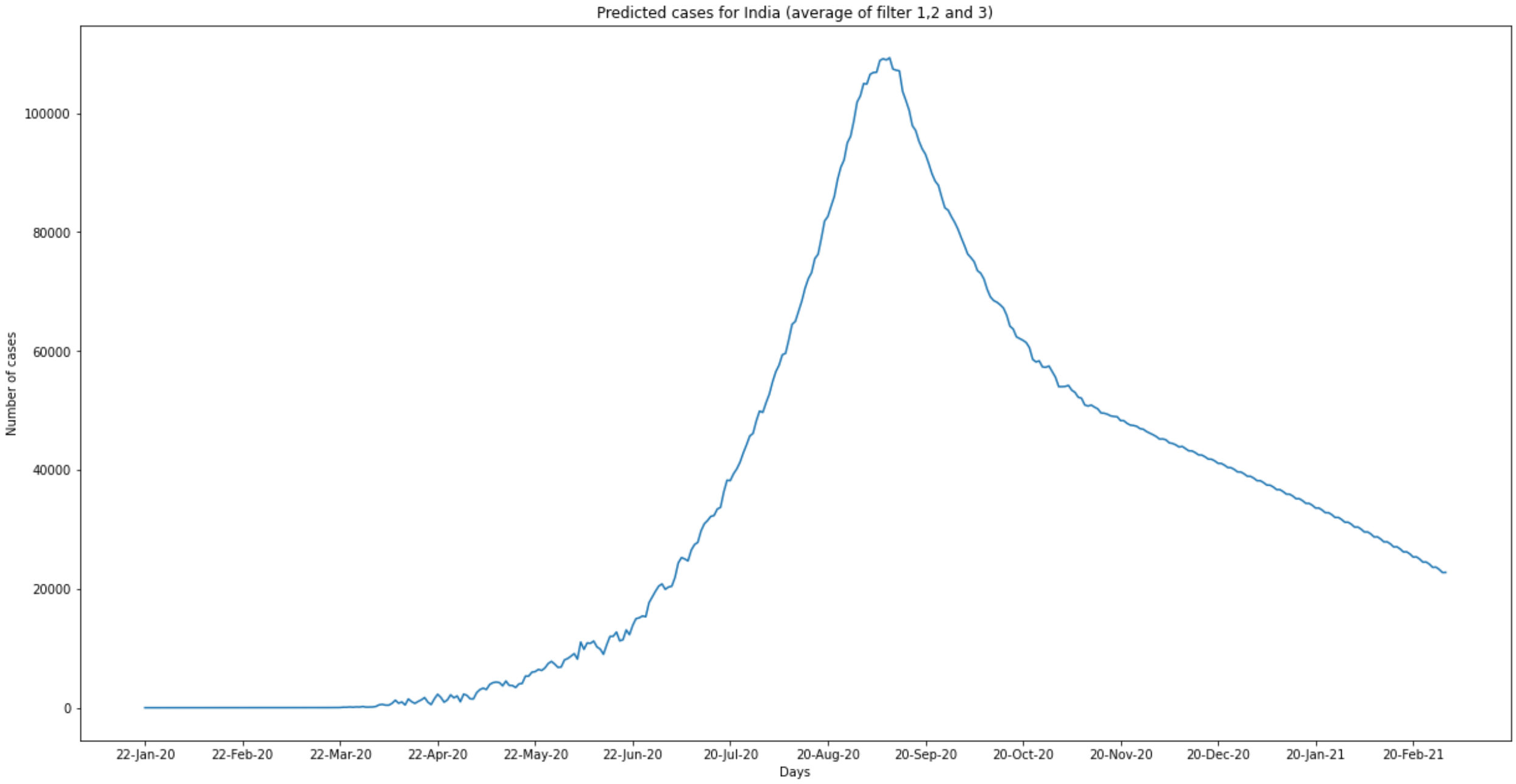

**Fig. 5** The average curve for India by using all the filters

## Conflicts of interest

On behalf of all authors, the corresponding author states that there is no conflict of interest. This article does not contain any studies with animals or Humans performed by any of the authors. All the necessary permissions were obtained from the Institute Ethical Committee and concerned authorities. No informed consent was required as the studies does not involve any human participant. No funding was involved in the present work.